\documentclass[journal]{IEEEtran}
\IEEEoverridecommandlockouts
\usepackage{cite}
\usepackage{amsmath,amssymb,amsfonts}
\usepackage{graphicx}
\usepackage{textcomp}
\usepackage{xcolor}
\def\BibTeX{{\rm B\kern-.05em{\sc i\kern-.025em b}\kern-.08em
    T\kern-.1667em\lower.7ex\hbox{E}\kern-.125emX}}

\definecolor{red_boston}{RGB}{204,0,0}
\usepackage{listings}
\usepackage{booktabs}
\usepackage{hyperref}
\usepackage{amsmath}
\usepackage{booktabs}
\usepackage{array}
\usepackage{balance}
\usepackage{algorithm}
\usepackage[]{algpseudocode}
    
\definecolor{codegreen}{rgb}{0,0.6,0}
\definecolor{codegray}{rgb}{0.5,0.5,0.5}
\definecolor{codepurple}{rgb}{0.58,0,0.82}
\definecolor{backcolour}{rgb}{0.95,0.95,0.92}
 
\lstdefinestyle{mystyle}{
    backgroundcolor=\color{backcolour},   
    commentstyle=\color{codegreen},
    keywordstyle=\color{magenta},
    numberstyle=\tiny\color{codegray},
    stringstyle=\color{codepurple},
    basicstyle=\footnotesize,
    breakatwhitespace=false,         
    breaklines=true,  
    postbreak=\mbox{\textcolor{red}{$\hookrightarrow$}\space},
    captionpos=b,                    
    keepspaces=true,                 
    numbers=left,                    
    numbersep=5pt,                  
    showspaces=false,                
    showstringspaces=false,
    showtabs=false,                  
    tabsize=2
}
 
\begin{document}

\title{The UCR Time Series Archive\\
}



\author{
Hoang Anh Dau, 
Anthony Bagnall,
Kaveh Kamgar, 
Chin-Chia Michael Yeh,
Yan Zhu, \\
Shaghayegh Gharghabi,
Chotirat Ann Ratanamahatana,
Eamonn Keogh
\thanks{H. A. Dau is with the Department of Computer Science and Engineering, University of California, Riverside,
CA, USA (e-mail: hdau001@ucr.edu).}
\thanks{A. Bagnall is with School of Computing Sciences, University of East Anglia, Norfolk, UK (email: ajb@uea.ac.uk).}
\thanks{K. Kamgar, C-C. M. Yeh, Y. Zhu, S. Gharghabi are with the Department of Computer Science and Engineering, University of California, Riverside,
CA, USA (e-mail: kkamg001@ucr.edu; myeh003@ucr.edu; yzhu015@ucr.edu; sghar003@ucr.edu).}
\thanks{C. A. Ratanamahatana is with Department of Computer Engineering, Chulalongkorn University, Bangkok, Thailand (email: chotirat.r@chula.ac.th).}
\thanks{E. Keogh is with the Department of Computer Science and Engineering, University of California, Riverside,
CA, USA (e-mail: eamonn@cs.ucr.edu).}
\thanks{Digital Object Identifier arXiv:1810.07758}
}

\maketitle

\begin{abstract}
The UCR Time Series Archive - introduced in 2002, has become an important resource in the time series data mining community, with at least one thousand published papers making use of at least one data set from the archive. The original incarnation of the archive had sixteen data sets but since that time, it has gone through periodic expansions. The last expansion took place in the summer of 2015 when the archive grew from 45 to 85 data sets. This paper introduces and will focus on the new data expansion from 85 to 128 data sets. Beyond expanding this valuable resource, this paper offers pragmatic advice to anyone who may wish to evaluate a new algorithm on the archive. Finally, this paper makes a novel and yet actionable claim: of the hundreds of papers that show an improvement over the standard baseline (1-nearest neighbor classification), a large fraction may be mis-attributing the reasons for their improvement. Moreover, they may have been able to achieve the same improvement with a much simpler modification, requiring just a single line of code.  
\end{abstract}

\IEEEpeerreviewmaketitle

\begin{IEEEkeywords}
Data mining, UCR time series archive, time series classification
\end{IEEEkeywords}

\section{Introduction}  

The discipline of time series data mining dates back to at least the early 1990s \cite{Agrawal1993}. As noted in a survey\cite{keogh2003need}, during the first decade of research, the vast majority of papers tested only on a single artificial data set created by the proposing authors themselves \cite{Agrawal1993, Huang1999, Kim2000, Saito1995}. While this is forgivable given the difficulty of obtaining data in the early days of the web, it made gauging progress and the comparisons of rival approaches essentially impossible. Frustrated by this difficulty \cite{keogh2003need}, and inspired by the positive contributions of the more general UCI Archive to the machine learning community \cite{Lichman2013}, Keogh \& Folias introduced the UCR Archive in 2002 \cite{Keogh2002}. The last expansion took place in 2015, bringing the number of the data sets in the archive to 85 data sets \cite{Chen2015}. As of Fall 2018, the archive has about 850 citations, but perhaps twice that number of papers use some fractions of the data set unacknowledged\footnote{Why would someone use the archive and not acknowledge it? \textit{Carelessness} probably explains the majority of such omissions. In addition, for several years (approximately 2006 to 2011), access to the archive was conditional on informally pledging to test on \textit{all} data sets to avoid cherry picking (see Section \ref{cherry-picking}). Some authors who did then go on to test on only a limited subset, possibly choosing not to cite the archive to avoid bringing attention to their failure to live up to their implied pledge.}. 

While the archive is heavily used, it has invited criticisms, both in published papers \cite{Hu2016} and in informal communications to the lead archivists (i.e. the current authors). Some of these criticisms are clearly warranted, and the 2018 expansion of the archive that accompanies this paper is designed to address some of the issues pointed out by the community. In addition, we feel that some of the criticisms are unwarranted, or at least explainable. We take advantage of this opportunity to, for the first time, explain some of the rationale and design choices made in producing the original archive. 

The rest of this paper is organized as follows. In Section \ref{baseline} we explain how the baseline accuracy that accompanies the archive is set. Section \ref{criticism} enumerates the major criticisms of the archive and discusses our defense or how we have addressed the criticisms with this expansion. In Section \ref{cherry-picking}, we demonstrate how bad the practice of ``cherry picking" can be, allowing very poor ideas to appear promising. In Section \ref{best-practices} we outline our best suggested practices for using the archive to produce forceful classification experiments. Section \ref{new-archive} introduces the new archive expansion. Finally, in Section \ref{conclusions} we summarize our contributions and provide directions for future work. 

\section{Setting the Baseline Accuracy \label{baseline}}

From the first iteration, the UCR Archive has had a single predefined train/test split, and three baseline (``strawman") scores accompany it. The baseline accuracies are from the classification result of the 1-Nearest Neighbor classifier (1-NN). Each test exemplar is assigned the class label of its closest match in the training set. The notion of ``closest match" is how similar the time series are under some distance measures. This is straightforward for Euclidean distance (ED), in which the data points of two time series are linearly mapped $i^{th}$ value to $i^{th}$ value. However, in the case of the Dynamic Time Warping distance (DTW), the distance can be different for each setting of the warping window width, known as the warping constraint parameter $w$ \cite{Dau2018LearningW}. 

\begin{figure}[htb]
\centering
\includegraphics[page=2,trim={10cm 5.4cm 10cm 5.4cm}, clip, width=1\columnwidth]{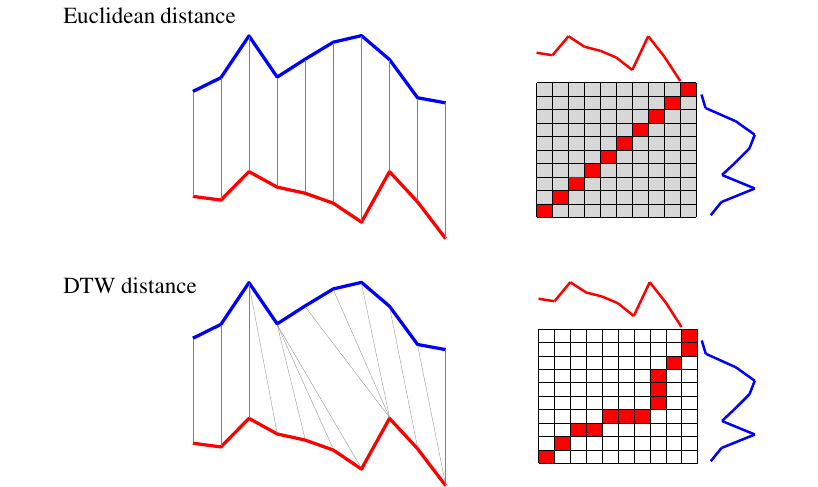}
\caption{
Visualization of the warping path. \textit{top}) Euclidean distance with one-to-one point matching. The warping path is strictly diagonal (cannot visit the grayed-out cells). \textit{bottom}) unconstrained DTW with one-to-many point matching. The warping path can monotonically advance through any cell of the distance matrix.
}
\label{fig:UCRArchive-p2}
\end{figure}

DTW allows non-linear mapping between time series data points. The parameter $w$ controls the maximum lead/lag for which points can be mapped to, thus preventing pathological mapping between two time series. The data points of two time series can be mapped $i^{th}$ value to $j^{th}$ value, with $|i -j| \leq s$, where $s$ is some integers, typically a small fraction of the time series length. In practice, this parameter is usually expressed as a percentage of the time series length and therefore, having values between 0 - 100\%. The use of DTW with $w = 100\%$ is called DTW with no warping window, or unconstrained DTW. The special case of DTW with $w = 0\%$ degenerates to the ED distance. This is illustrated in Fig.~\ref{fig:UCRArchive-p2}.

The setting of $w$ can have a significant effect to the clustering and classification result \cite{Dau2018LearningW}. If not set carefully, a poor choice for this parameter can drastically deteriorate the classification accuracy. For most problems, a $w$ greater than 20\% is not needed and likely only imposes a computational burden.
 
We refer to the practice of using 1-NN with Euclidean distance as 1-NN ED, and the practice of using 1-NN with DTW distance as 1-NN DTW. The UCR Time Series Archive reports three baseline classification results. These are classification error rate of:

\begin{itemize}
    \item 
	1-NN Euclidean distance
	\item
	1-NN unconstrained DTW 
	\item
	1-NN constrained DTW with learned warping window width
\end{itemize}

\begin{figure}[htb]
\centering
\includegraphics[page=3,trim={10cm 5.3cm 10cm 5.3cm}, clip, width=0.95\columnwidth]{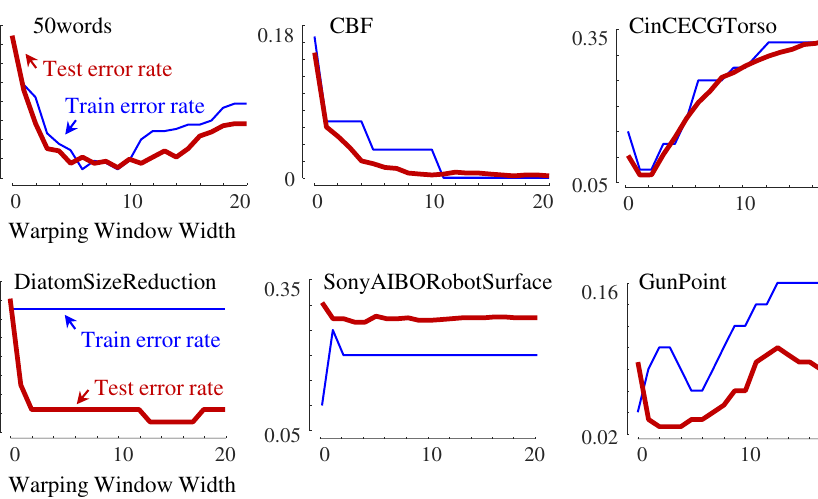}
\caption{
\textit{\textcolor{blue}{blue}/fine}) The leave-one-out error rate for increasing values of warping window $w$, using DTW-based 1-nearest neighbor classifier. \textit{\textcolor{red_boston}{red}/bold}) The holdout error rate. In the bottom-row examples, the holdout accuracies do not track the predicted accuracies.
}
\label{fig:UCRArchive-p3}
\end{figure}

For the last case, we must \textit{learn} a parameter from the training data. The best warping window width is decided by performing Leave-One-Out Cross-Validation (LOO CV) with the train set, choosing the smallest value of $w$ that minimizes the average train error rate. Generally, this approach works well in practice. However, it can produce poor results as in some situations, the best $w$ in training may not be the best $w$ for testing. The top row of Fig.~\ref{fig:UCRArchive-p3} shows some examples where the learned constraint closely predicts the effect the warping window will have on the unseen data. The bottom row of Fig.~\ref{fig:UCRArchive-p3}, in contrast, shows some examples where the learned constraint fails to track the real test error rate, thus giving non-optimal classification result on holdout data.

Happily, the former case is much more common \cite{Dau2018LearningW}. When does learning the parameter fail? Empirically, the problem only occurs for very small training sets; however, this issue is common in real world deployments.    

\section{Criticisms of the UCR Archive \label{criticism}}

In this section we consider the criticisms that have been levelled at the UCR Archive. We enumerate and discuss them in no particular order.

\subsection{Unrealistic Assumptions}

Bing et al. have criticized the archive for the following unrealistic assumptions \cite{Hu2016}.
\begin{itemize}
    \item 
	\textit{There is a copious amount of perfectly aligned atomic patterns.} However, in at least in some domains, labeled training data can be expensive or difficult to obtain.
	\item
	\textit{The patterns are all of equal length.} In practice, many patterns reflecting the same behavior can be manifest at different lengths. For example, a natural walking gait cycle can vary by at least plus or minus 10\% in time.
	\item
	\textit{Every item in the archive belongs to exactly one well-defined class; there is no option to choose an \texttt{``unknown"} or \texttt{``unclassifiable"}.} For example, in the \textit{Cricket} data sets, each signal belongs to one of the twelve classes, representing the hand signs made by an umpire. However, for perhaps 99\% of a game, the umpire is not making \textit{any} signal. It cafn be argued that any practical system needs to have a thirteenth class named \texttt{``not-a-sign"}. This is not trivial, as this class will be highly variable, and this would create a skewed data set.  
\end{itemize}

\subsection{The Provenance of the Data is Poor}

Here we can only respond \textit{mea culpa}. The archive was first created as a small-scale personal project for Keogh's lab at University of California, Riverside. We did not know at the time that it would expand so large and become an important resource for the community. In this release, we attempt to document the data sets in a more systematic manner. In fact, one of the criteria for including a new data set in the archive is that it has a detailed description from the data donor or it has been published in a research paper that we could cite. 

\subsection{Data Already Normalized}

The time series are already z-normalized to remove offset and scaling (transformed data have zero mean and in unit of standard deviation). The rationale for this step was previously discussed in the literature \cite{Rakthanmanon2013}; we will briefly review it here with an intuitive example. 

Consider the \textit{GunPoint} data set shown in Fig.~\ref{fig:UCRArchive-p8}. Suppose that we did not z-normalize the data but allowed our classifier to exploit information about the exact \textit{absolute} height of the gun or hand. As it happens, this \textit{would} help a little. However, imagine we collected more test data next week. Further suppose that for this second session, the camera zoomed in or out, or the actors stood a little closer to the camera, or that the female actor decided to wear new shoes with a high heel. None of these differences would affect z-normalized data as z-normalization accounts for offset and scale variance; however, they would drastically (negatively) affect any algorithm that exploited the raw un-normalized values. 

Nevertheless, we acknowledge that for some (we believe, \textit{very} rare) cases, data normalization is ill-advised. For the new data sets in this release, we provide the raw data without any normalization when possible; we explicitly state if the data has been normalized beforehand by the donors (the data might have been previously normalized by the donating source, who lost access to original raw data).

\subsection{The Individual Data Sets are Too Small}

While it is true that there is a need for bigger data sets in the era of ``big data" (some algorithms specifically target scaling for big data sets), the archive has catered a wide array of data mining needs and lived up to its intended scope. The largest data set is \textit{StarLightCurves} with 1,000 train and 8,236 test objects, covering 3 classes. The smallest data set is \textit{Beef} with 30 train and 30 test objects, covering 5 different classes. Note that in recent years, there have been several published papers that state something to the effect of \textit{``in the interests of time, we only tested on a subset of the archive".} Perhaps a specialist archive of massive time series can be made available for the community in a different repository.

\subsection{The Data Sets are Not Reflective of Real-world Problems} 
This criticism is somewhat warranted. The archive is biased towards:
\begin{itemize}
    \item 
	data sets that reflect the personal interests/hobbies of the principal investigator (PI), Eamonn Keogh, including entomology (\textit{InsectWingbeatSound}), anthropology (\textit{ArrowHead}) and astronomy (\textit{StarLightCurves}). A wave of data sets added in 2015 reflect the personal and research interests of Tony Bagnall \cite{Bagnall2018}, many of which are image-to-time-series data sets. The archive has always had a policy of adding \textit{any} donated data set, but offers of donations are surprisingly rare. Even when we actively solicited donations by writing to authors and asking for their data, we found that only a small subset of authors is willing to share data. The good news is that there appears to be an increasing willingness to share data, perhaps thanks to conferences and journals actively encouraging reproducible research. 
	\item
	data sets that could be easily obtained or created. For example, fMRI data could be very interesting to study, but the PI did not have access to such a machine or the domain knowledge to create a classification data set in this domain. However, with an inexpensive scanner or a camera, it was possible to create many image-derived data sets such as \textit{GunPoint, OSULeaf, SwedishLeaf, Yoga, Fish} or \textit{FacesUCR}.
	\item
	data sets that do not have privacy issues. For many domains, mining the data while respecting privacy is an important issue. Unfortunately, none of the data sets in the UCR Archive motivates the need for privacy (though it is possible to use the data to construct proxy data sets).
\end{itemize}

\subsection{Benchmark Results are from a Single Train/Test Split} \label{singleTrainTest}

Many researchers, especially those coming from a traditional machine learning background have criticized the archive for having a single train/test split. The original motivation for fixing the train and test set was to allow \textit{exact} reproducibility. Suppose we simply suggested doing five-fold cross validation. Further suppose, someone claimed to be able to achieve an accuracy of $A$, on some data sets in the archive. If someone else re-implemented their algorithm and got an accuracy that is slightly lower than $A$ during their five-fold cross validation, it would be difficult to know if that was within the expected variance of different folds, or the result of a bug or a misunderstanding in the new implementation. This issue would be less of a problem if everyone shared their code, and/or had very explicit algorithm descriptions. However, while the culture of open source code is growing in the community, such openness was not always the norm. 

With a single train/test split, and a deterministic algorithm such as 1-NN, failure to \textit{exactly} reproduce someone else`s result immediately suggests an issue that should be investigated before proceeding with research. Note that while performing experiments on the single train/test split was always suggested as an absolute minimum sanity check; it did/does not preclude pooling the two splits and then performing $K$-fold cross validation or any other more rigorous evaluation. 

\section{How Bad is Cherry Picking? \label{cherry-picking}}

It is not uncommon to see papers which report only results on a subset of the UCR Archive, without any justification or explanation. Here are some examples. 
\begin{itemize}
    \item 
	\textit{``We evaluated 1D-SAXLSSS classification accuracy on 22 data sets (see Table 2) taken from publicly available UCR repository benchmark"} \cite{Taktak2017}  
	\item
	\textit{``Figure 3 shows a performance gain of DSP-Class-SVM and DSP-Class-C5.0 approach in 5/11 data sets compared to another technique that does not use features (1NN with Euclidean distance)"} \cite{Silva2016a}. 
	\item
	\textit{``We experiment 48 small-scale data sets out of total 85 problems in the UCR time series archive"} \cite{He2018}
\end{itemize}

We have no way to determine if these authors cherry-picked their limited subset of the archive, they may have selected the data on some unstated whim that has nothing to do with classification accuracy. However, without a statement of what that whim might be, we cannot exclude the possibility. Here, we will show how cherry picking can make a vacuous idea look good. Again, to be clear we are not suggesting that the works considered above are in any way disingenuous.  

Consider the following section of text (italicized for clarity) with its accompanying table and figure, and imagine it appears in a published report. While this is a fictional report, note that all the numbers presented in the table and figure are \textit{true} values, based on reproducible experiments that we performed \cite{Dau2019UCRArchivePaperWebpage}. 

\textit{We tested our novel FQT algorithm on 20 data sets from the UCR Archive. We compared to the Euclidean distance, a standard benchmark in this domain. Table T summarizes the results numerically, and Fig.~F shows a scatter plot visualization.} 

\textit{Table T: Performance comparison between Euclidean distance and our FQT distance. Our proposed FQT distance wins on all data sets that we consider.}
\vspace{0.5cm}
\vspace{0.1cm}
\texttt{
\scriptsize
    \begin{tabular}{ l m{0.8cm} m{0.8cm} m{1.1cm} }  
        \toprule
        data set & ED Error & FQT Error & Error Reduction  \\
        \midrule
        Strawberry  &   0.062   &	0.054   &	0.008   \\
        ECG200  &	0.120   &	0.110   &	0.010   \\
        TwoLeadECG  &	0.253   &	0.241   &	0.012   \\
        Adiac   &	0.389   &	0.376   &	0.013   \\
        ProximalPhalanxTW   &	0.292   &	0.278   &	0.014   \\
        DistalPhalanxTW &	0.273   &	0.258   &	0.015   \\
        ProximalPhalanxOutlineCorrect   &	0.192   &	0.175   &	0.017   \\
        RefrigerationDevices    &	0.605   &	0.587   &	0.018   \\
        Wine    &	0.389   &	0.370   &	0.019   \\
        ProximalPhalanxOutlineAgeGroup  &	0.215   &	0.195   &	0.020   \\
        Earthquakes &	0.326   &	0.301   &	0.025   \\
        ECGFiveDays &	0.203   &	0.177   &	0.026   \\
        SonyAIBORobotSurfaceII  &	0.141   &	0.115   &	0.026   \\
        Lightning7  &	0.425   &	0.397   &	0.028   \\
        Trace   &	0.240   &	0.210   &	0.030   \\
        MiddlePhalanxTW &	0.439   &	0.404   &	0.035   \\
        ChlorineConcentration   &	0.350   &	0.311   &	0.039   \\
        BirdChicken &	0.450   &	0.400   &	0.050   \\
        Herring &	0.484   &	0.422   &	0.062   \\
        CBF &	0.148   &	0.080   &	0.068   \\
        \bottomrule
    \end{tabular}
 }   
\vspace{0.5cm}

\textit{Note that we used identical (UCR pre-defined) splits for both approaches, and an identical classification algorithm. Thus, all improvements can be attributed to our novel distance measure. The improvements are sometimes small, however, for CBF, Herring and BirdChicken, they are 5\% (0.05) or greater, demonstrating that our FQT distance measure potentially offers significant gains in some domains. Moreover, we prove that FQT is a metric, and therefore easy to index with standard tree access methods.}

\begin{figure}[htb]
\centering
\includegraphics[page=4,trim={9.75cm 5cm 9.75cm 5cm}, clip, width=1\columnwidth]{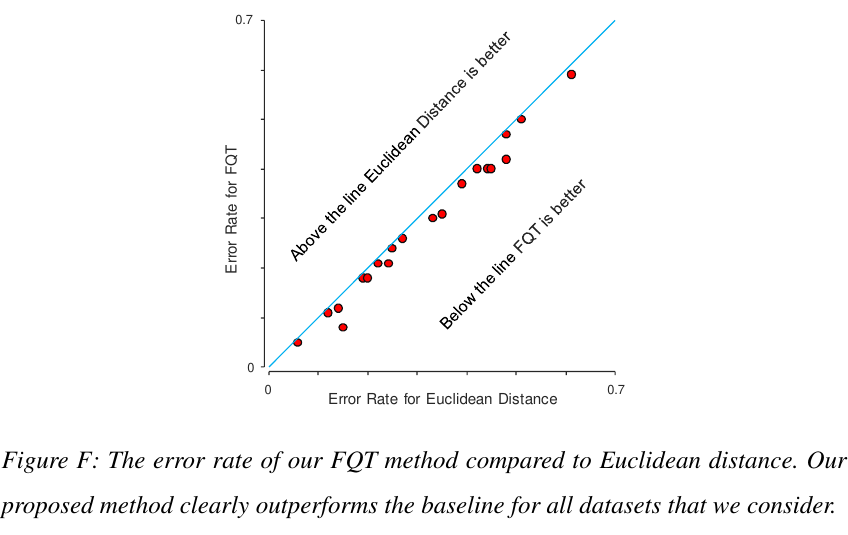}
\label{fig:UCRArchive-p4}
\end{figure}

(Returning to the current authors voice)
The above results are all true, and the authors are correct in saying that FQT is a metric and is easier to index than DTW. So, what is this remarkable FQT distance measure? It is simply the Euclidean distance after the first 25\% of each time series is thrown away (First Quarter Truncation). Here is how we compute the FQT distance for time series A and B in MATLAB:

\begin{lstlisting}[language=Matlab,numbers=none]
FQT_dist = sqrt(sum((A(end*0.25:end) - B(end*0.25:end)).^2))
\end{lstlisting}

If we examine the full 85 data sets in the archive, we will find that FQT wins on 19 data sets, but loses/draws on 66 (if we count a win as at least 1\% reduction in error rate). Moreover, the size of the losses is generally more dramatic than the wins. For instance, the ``error reduction" for \textit{MedicalImages} is -0.23 (accuracy decreases by 23\%). 

Simply deleting the first quarter of every time series is obviously not a clever thing to do, and evaluating this idea on all the data sets confirms that. However, by cherry picking the twenty data sets that we chose to report, we made it seem like very good idea. It is true that in a full paper based on FQT we would have had to explain the measure, and it would have struck a reader as simple, unprincipled and unlikely to be truly useful. However, there are many algorithms that would have the same basic \textit{``a lot worse on most, a little better on a few"} outcome, and many of these could be framed to sound like plausible contributions (cf. Section \ref{cautionary-tale}). 

In a recent paper, Lipton \& Steinhardt list some \textit{``troubling trends in machine learning scholarship''} \cite{Lipton2018}. One issue identified is ``mathiness", defined as \textit{``the use of mathematics that obfuscates or impresses rather than clarifies"}. We have little doubt that we could ``dress up" our proposed FQT algorithm with spurious notation (Lipton and Steinhardt \cite{Lipton2018} call it \textit{``fancy mathematics"}) to make it sound complex. 

To summarize this section, cherry picking can make an arbitrary poor idea look useful, or even wonderful. Clearly, not all (or possibly even, not \textit{any}) papers that report on a subset of the UCR data sets are trying to deceive the reader. However, as an outsider to the research effort, it is essentially impossible to know if the subset selection was random and fair (made \textit{before} any results were computed) or biased to make the approach appear better than it really is.

The reasons given for testing only on a subset of the data (where any reason is given at all) is typically something like \textit{``due to space limitations, we report only five of the ..."}. However, this is not justified. A Critical Difference Diagram like Fig.~\ref{fig:UCRArchive-p6} or a scatter plot like Fig.~\ref{fig:UCRArchive-p7} require very little space but can summarize an arbitrary number of data sets. Moreover, one can always place detailed spreadsheets online or in an accompanying, cited technical report, as many papers do these days \cite{Paparrizos2015, Dau2018LearningW}.

That being said, we believe that sometimes there are good reasons to test a new algorithm on only a subset of the archive, and we applause researchers who explicitly justify their conduct. For example, Hills et al. stated: \textit{``We perform experiments on 17 data sets from the UCR time-series repository. We selected these particular UCR data sets because they have relatively few cases; even with optimization, the shapelet algorithm is time consuming."} \cite{Hills2014}.

\section{Best Practices for Using the Archive \label{best-practices}}

\begin{figure}[htb]
\centering
\includegraphics[page=5,trim={8.5cm 6.8cm 8.5cm 6.8cm}, clip, width=1\columnwidth]{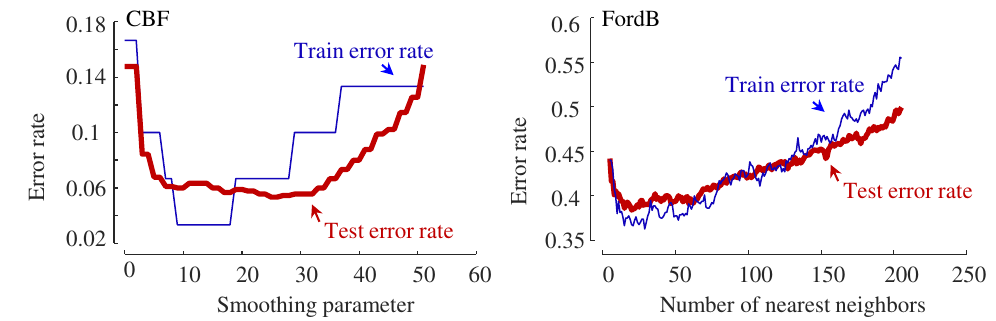}
\caption{
\textit{left}) The error rate on classification of the \textit{CBF} data set for increasing amounts of smoothing using MATLAB's default smoothing algorithm. \textit{right}) The error rate on classification of the \textit{FordB} data set for increasing number of nearest neighbors. Note that the leave-one-out error rate on the training data does approximately predict the best parameter to use. 
}
\label{fig:UCRArchive-p5}
\end{figure}

Beating the performance of DTW on some data sets should be considered a \textit{necessary}, but not \textit{sufficient} condition for introducing a new distance measure or classification algorithm. This is because the performance of DTW itself can be improved with very little effort, in at least a dozen ways. In many cases, these simple improvements can close most or all the gap between DTW and the more complex measures being proposed. For example:
\begin{itemize}
    \item 
	The warping window width parameter of constrained DTW algorithm is tuned by the ``quick and dirty" method described in Section \ref{baseline}. As Fig.~\ref{fig:UCRArchive-p3} bottom row shows, on at least some data sets, that tuning is sub-optimal. The parameter could be tuned more carefully in several ways such as by re-sampling or by creating synthetic examples \cite{Dau2017}.
	\item
	The performance of DTW classification can often be improved by other trivial changes. For example, as shown in Fig.~\ref{fig:UCRArchive-p5}.left, simply smoothing the data can produce significant improvements. Fig.~\ref{fig:UCRArchive-p5}.right shows that generalizing from 1-nearest neighbor to k-nearest neighbor often helps. One can also test alternative DTW step patterns \cite{Lu2017}. Making DTW ``endpoint invariant" helps on many data sets \cite{Silva2017}, etc.
\end{itemize}

An hour spent on optimizing any of the above could improve the performance of ED/DTW on at least several data sets of the archive. 

\subsection{Mis-attribution of Improvements: a Cautionary Tale \label{cautionary-tale}}

We believe that of the several hundred papers that show an improvement on the baselines for the UCR Archive, a not small fraction is mis-attributing the cause of their improvement and is perhaps \textit{indirectly} discovering one of the low hanging fruits above. This point has recently been made in the more general case by researchers who believe that many papers suffer from a failure \textit{``to identify the sources of empirical gains"} \cite{Lipton2018}. Below we show an example to demonstrate this.

Many papers have suggested using a wavelet representation for time series classification\footnote{These works should not be confused with papers that suggest using a wavelet representation to perform dimensionality reduction to allow more efficient indexing of time series.}, and have gone on to show accuracy improvements over either the DTW or ED baselines. In most cases, these authors attribute the improvements to the multi-resolution properties of wavelets. For example (our emphasis in the quotes below):
\begin{itemize}
    \item 
	\textit{``wavelet compression techniques can sometimes even help achieve higher classification accuracy than the raw time series data, as they better capture essential local features ... As a result, we think it is safe to claim that \underline{multi-level} wavelet transformation is indeed helpful for time series classification."} \cite{Li2016}
	\item
	\textit{``our \underline{multi-resolution} approach as discrete wavelet transforms have the ability of reflecting the local and global information content at every resolution level."} \cite{Li2016}
	\item
	\textit{``We attack above two problems by exploiting the \underline{multi-scale} property of wavelet decomposition ... extracting features combining the global information and partial information of time series."} \cite{Zhang2006}
	\item
	\textit{``Thus \underline{multi-scale} analyses give us the ability of observing time series in various views."} \cite{Zhang2005}
\end{itemize}

As the quotes above suggest, many authors attribute their accuracy improvements to the multi-resolution nature of wavelets. However, we have a different hypothesis. The wavelet representation is simply smoothing the data implicitly, and all the improvements can be attributed to \textit{just} this smoothing! Fig.~\ref{fig:UCRArchive-p5} above does offer evidence that at least on some data sets, appropriate smoothing is enough to make a difference that is commensurate with the claimed improvements. However, is there a way in which we could be sure? Yes, we can exploit an interesting property of the Haar Discrete  Wavelet  Transform (DWT).

Note that if both the original and reduced dimensionality of the Haar wavelet transform are powers of two integers, then the approximation produced by Haar is logically \textit{identical} to the approximation produced by the Piecewise Aggerate Approximation \cite{Keogh2001}. This means that the distances calculated in the truncated coefficient space are identical for both approaches, and thus they will have the same classification predictions \textit{and} the same error rate. If we revisit the \textit{CBF} data set shown in Fig.~\ref{fig:UCRArchive-p5}, using Haar with 32 coefficients we get an error rate of just 0.05, much better than the 0.148 we would have obtained using the raw data. Critically, however, PAA with 32 coefficients also gets the same 0.05 error rate.

It is important to note that PAA is not in any sense \textit{multi-resolution} or \textit{multiscale}. Moreover, by the definition of PAA, each coefficient being the \textit{average} of a \textit{range} of points, is very similar to the definition of the moving average filter smoothing, with each point being \textit{averaged} with its neighbors within a \textit{range} to the left and the right.  
We can do one more test to see if the Haar Wavelet is offering us something beyond smoothing. Suppose we use it to classify data that have \textit{already} been smoothed with a smoothing parameter of 32. Recall (Fig.~\ref{fig:UCRArchive-p5}) that using this smoothed data directly gives us an error rate of 0.055. Will Haar Wavelet classification further improve this result? No, in fact using 32 coefficients, both Haar Wavelet and PAA classification produce very slightly worse results of 0.057, presumably because we have effectively smoothed the data twice, and by doing so, over-smoothed it (again, see Fig.~\ref{fig:UCRArchive-p5}, and examine the trend of the curve as the smoothing parameter grows above 30). In our view, these observations cast significant doubt on the claim that the improvements obtained can correctly be attributed to multi-resolution properties of wavelets. 

There are two obvious reasons as to why this matters:
\begin{itemize}
    \item 
	mis-attribution of \textit{why} a new approach works potentially leaves adopters in the position of fruitless follow-up work or application of the proposed ideas to data/tasks for which they are not suited;
	\item
	if the problem at hand is really to improve the accuracy of time series classification, and if five minutes spent experimenting with a smoothing function can give you the same improvement as months implementing a more complex method, then surely the former is more desirable, only less publishable.  
\end{itemize}

This is simply one concrete example. We suspect that there are many other examples. For instance, many papers have attributed time series classification success to their exploitation of the ``memory" of Hidden Markov Models, or ``long-term dependency features" of Convolution Neural Networks etc. However, one almost never sees an ablation study that forcefully convinces the reader that the \textit{claimed} reason for improvement is correct. 

In fairness, a handful of papers do explicitly acknowledge that. While they may introduce a complex representation or distance measure for classification, at least some of the improvements should be attributed to smoothing. For example, Sch\"afer notes: \textit{``Our Bag-of-SFA-Symbols (BOSS) model combines the extraction of substructures with the tolerance to extraneous and erroneous data using a noise reducing representation of the time series"} \cite{Schafer2015}. Likewise, Li and colleagues \cite{Li2016a} revisit their Discrete Wavelet Transformed (DWT) time series classification work \cite{Li2016} to explicitly ask \textit{``if the good performances of DWT on time series data is due to the implicit smoothing effect"}. They show that their previous embrace of wavelet-based classification does not produce results that are better than simple smoothing in a statistically significant way. Such papers, however, remain an exception.

\subsection{How to Compare Classifiers}

\subsubsection{The choice of performance metric}

Suppose you have an archive of one hundred data sets and you want to test whether classifier A is better than classifier B, or compare a set of classifiers, over these data. At first, you need to specify what you mean by one classifier being ``better" than another. There are two general criteria with which we may wish compare classifier ability on data not used in the training process: the prediction ability and the probability estimates. 

The ability to predict is most commonly measured by accuracy (or equivalently, error rate). However, accuracy does not always tell the whole story. If a problem has class imbalance, then accuracy may be less informative than a measure that compensates for this skewed classes. Sensitivity, specificity, precision, recall and the F statistic are all commonly used for two-class problems where one class is rarer than the other, such as medical diagnosis trials \cite{Okeh2012, Uguroglu2013}. However, these measures do not generalize well to multi-class problems or scenarios where we cannot prioritize one class using domain knowledge. For the general case over multiple diverse data sets, we consider accuracy and balanced accuracy enough to assess predictive power. Conventional accuracy is the proportion of examples correctly classified while balanced accuracy is the average of accuracy for each class individually \cite{Brodersen2010}. 
Some classifiers produce scores or probability estimates for each class and these can be summarized with statistics such as negative log-likelihood or area under the receiver operating characteristic curve (AUC). However, if we are using a classifier that only produces predictions (such as 1-NN), these metrics do not apply.

\subsubsection{The choice of data split}

Having decided on a comparison metric, the next issue is what data you are going to evaluate the data on. If a train/test split is provided, it is natural to start by building all the classifiers (including all model selection/parameter setting) on the train data, and then assess accuracy on the test data. 

There are two main problems with using a single train test split to evaluate classifiers. First, there is a temptation to cheat by setting the parameters to optimize the test data. This can be explicit, for example, by setting an algorithm to stop learning when test accuracy is maximized, or implicit, by setting default values based on knowledge of the test split. For example, suppose we have generated results such as Fig.~\ref{fig:UCRArchive-p5}, and have a variant of DTW we wish to assess on this test data. We may perform some smoothing and set the parameter to a default of 10. Explicit bias can only really be overcome with complete code transparency. For this reason, we \textit{strongly} encourage users of the archive to make their code available to both reviewers and readers.

The other problem with a single train/test split, particularly with small data sets, is that tiny differences in performance can seem magnified. For example, we were recently contacted by a researcher who queried our published results for 1-NN DTW on the UCR archive train/test splits. When comparing our accuracy results to theirs, they noticed that they differ by as much as 6\% in some instances, but there was no significant difference for other problem sets. Still, we were concerned by this single difference, as the algorithm in question is deterministic. On further investigation, we found out that our data were rounded to six decimal places, theirs to eight. These differences on single splits were caused by small data set sizes and tiny numerical differences (often \textit{just} a single case classified differently). 

These problems can be largely overcome by merging the train and test data and re-sampling each data set multiple times then averaging test accuracy. If this is done, there are several caveats: 
\begin{itemize}
    \item 
	the default train and test splits should always be included as the first re-sample;
	\item
	re-samples must be the same for each algorithm; 
	\item
	the re-samples should retain the initial train and test sizes;
	\item
	the re-samples should be stratified to retain the same class distribution as the original. 
\end{itemize}

Even when meeting all these constraints, re-sampling is not always appropriate. Some data are constructed to keep experimental units of observation in difference data sets. For example, when constructing the alcohol fraud detection problem \cite{Lines2016}, we used different bottles in the experiments and made sure that observations from the same bottle does not appear in both the train and the test data. We do this to make sure we are not detecting bottle differences rather than different alcohol levels. Problems such as these discourage practitioners from re-sampling. However, we note that the majority of machine learning research involves repeated re-samples of the data.

Finally, for clarity we will repeat our explanation in Section~\ref{singleTrainTest} as to why we have a single train/test split in the archive. Publishing the results of both ED and DTW distance on a single deterministic split provides researchers a useful sanity check, before they perform more sophisticated analysis \cite{Demsar2006, garcia2008extension, Salzberg1997}.

\subsubsection{The choice of significance tests} \label{significanceTest}

Whether through a single train/test split or through re-sampling then averaging, you now arrive at a position of having multiple accuracy estimates for each classifier. The core question is, are there significant differences between the classifiers? In the simpler scenario, suppose we have two classifiers, and we want to test whether the differences in average accuracy is different from zero. There are two alternative hypothesis tests that one could go for, a paired two-sample t-test \cite{Student1908} for evidence of a significant difference in mean accuracies or a Wilcoxon signed-rank test \cite{Wilcoxon1945, Siegal1956} for differences in median accuracies. Generally, machine learning researchers favor the latter. However, it is worth noting that many of the problems identified with parametric tests in machine learning derive from the problem of too few data sets, typically twenty or less. With more than 30 data sets, the central limit theorem means these problems are minimized. Nevertheless, we advise using a Wilcoxon signed-rank test with a significance level of $\alpha$ set at or smaller than 0.05 to satisfy reviewers. 

\begin{figure}[htb]
\centering
\includegraphics[page=6,trim={11.5cm 7.8cm 11.5cm 7.8cm}, clip, width=0.95\columnwidth]{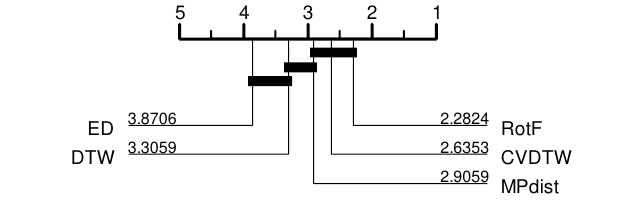}
\caption{
Critical difference for MPdist distance against four benchmark distances. Figure credited to Gharghabi et al. \cite{Gharghabi2018}. We can summarize this diagram as follow: RotF is the best performing algorithm with an average rank of 2.2824; there is an overall significant difference among the five algorithms; there are three distinct cliques; MPdist is significantly better than ED distance and not significantly worse than the rest.
}
\label{fig:UCRArchive-p6}
\end{figure}

What if you want to compare multiple classifiers on these data sets? We follow the recommendation of Dem\v{s}ar \cite{Demsar2006} and base the comparison on the ranks of the classifiers on each data set rather than the actual accuracy values. We use the Friedmann test \cite{Friedman1937, Friedman1940} to determine if there were any statistically significant differences in the rankings of the classifiers. If differences exist, the next task is to determine where they lie. This is done by forming cliques, which are groups of classifiers within which manifest significant difference. Following recent recommendations in \cite{Benavoli2016} and \cite{garcia2008extension}, we have abandoned the Nemenyi post-hoc test \cite{hollander2013nonparametric} originally used by Dem\v{s}ar \cite{Demsar2006}. Instead, we compare all classifiers with pairwise Wilcoxon signed-rank tests and form cliques using the Holm correction, which adjusts family-wise error less conservatively than a Bonferonni adjustment \cite{Holm1979}. We can summarize these comparisons in a critical difference diagram such as Fig.~\ref{fig:UCRArchive-p6}.

Fig.~\ref{fig:UCRArchive-p6} displays the performance comparison between MPdist, a recently proposed distance measure and other competitors \cite{Gharghabi2018}. This diagram orders the algorithms and presents the average rank on the number line. Cliques of methods are grouped with a solid bar showing groups of methods within which there is no significant difference. According to Fig.~\ref{fig:UCRArchive-p6}, the best ranked method is Rotation Forest (RotF), however, it is not statistically better than the other two methods in its clique, CVDTW and MPdist.

\subsection{A Checklist}

We propose the following checklist for any researcher who is proposing a novel time series classification/clustering technique and would like to test it on the UCR Archive.
\begin{enumerate}
    \item 
	Did you test on \textit{all} the data sets? If not, you should carefully explain why not, to avoid the appearance of cherry picking. For example, \textit{``we did not test on the large data sets, because our method is slow"} or \textit{``our method is only designed for ECG data, so we only test on the relevant data sets"}.
	\item
	Did you take care to optimize all parameters on just the training set? Producing a Texas Sharpshooter plot is a good way to visually confirm this for the reviewers \cite{Batista2014}.
	Did you perform an appropriate statistical significance test? (see Section \ref{significanceTest})  
	\item
	If you are claiming your approach is better due to property X, did you conduct an ablation test (lesion study) to show that if you remove this property, the results worsen, and/or, if you endow an otherwise unrelated approach with this property, that approach also improves?
	\item
	Did you share your code? Note, some authors state in their submission, something to the effect of \textit{``We will share code when the paper gets accepted"}. However, sharing of documented code at the time of submission is the best way to imbue confidence in even the most cynical reviewers. 
	\item
	If you modified the data in any way (adding noise, smoothing, interpolating, etc.), did you share the modified data, or the code, with random seed generator, that would allow a reader to exactly reproduce the data. 
\end{enumerate}

\section{The New Archive \label{new-archive}}

On January 2018, we reached out about forty researchers soliciting ideas for the new UCR Time Series Archive. They are among the most active researchers in the time series data mining community, who have used the UCR Archive in the past. We raised the question: \textit{``What would you like to see in the new archive?"}. We saw a strong consensus on the following needs: 
\begin{itemize}
    \item 
	Longer time series
	\item
	Variable length data sets
	\item
	Multi-variate data sets
    \item
	Information about the provenance of the data sets 
\end{itemize}
Some researchers also wish to see the archive to include data sets suitable for some specific research problems. For example:
\begin{itemize}
    \item 
	data sets with highly unbalanced classes
	\item
	data sets with very small training set to benchmark data augmentation techniques
\end{itemize}

Researchers especially raised the need for bigger data sets in the era of big data.

\textit{``To me, the thing we lack the most is larger data sets; the largest training data has 8,000 time series while the community should probably move towards millions of examples. This is a wish, but this of course doesn't go without problems: how to host large data sets, will future researchers have to spend even more time running their algorithms, etc."} (Fran\c{c}ois Petitjean, Monash University)

Some researchers propose sharing pointers to genuine data repositories and data mining competitions. 

\textit{``A different idea that might be useful is to add data set directories that have pointers to other data sets that are commonly used and freely available. When the UCR Archive first appeared, it was a different time, with fewer high quality, freely available data sets that were used by many researchers to compare results. Today there are many such data sets, but you tend to find them with Google, or seeing mentions in papers. One idea would be to pull together links to those data sets in one location with a flexible ``show me data sets with these properties or like this data set" function."} (Tim Oates, University of Maryland Baltimore County).

While some of these ideas may go beyond the ambition of the UCR Archive, we think that it inspires a wave of effort in making the time series community better. We hope \textit{others} will follow suit our effort in making data sets available for research purposes. In a sense, we think it would be inappropriate for our group to provide all such needs, as this monopoly might bias the direction of research to reflect our interests and skills.  

We have addressed \textit{some} of the perceived problems with the existing archive and some of what the community want to see in the new archive. We follow with an introduction of the new archive release. 

\subsection{General Introduction}

We refer to archive before the Fall 2018 expansion as the old archive and the current version as the \textit{new} archive. The Fall 2018 expansion increases the number of data sets from 85 to 128. We adopt a standard naming convention, that is using captions for words and no underscores. Where possible, we include the provenance of the data sets. In editing data for the new archive, in most cases, we make the test set bigger than the train set to reflect real-world scenarios, i.e., labeled train data are usually expensive. We keep the data sets as is if they are from a published papers and are already in a UCR Archive preferred format (see guidelines for donating data sets to the archive \cite{Dau2018UCRArchivePage}).

\begin{figure}[htb]
\centering
\includegraphics[page=7,trim={8.9cm 2cm 8.9cm 2cm}, clip, width=1\columnwidth]{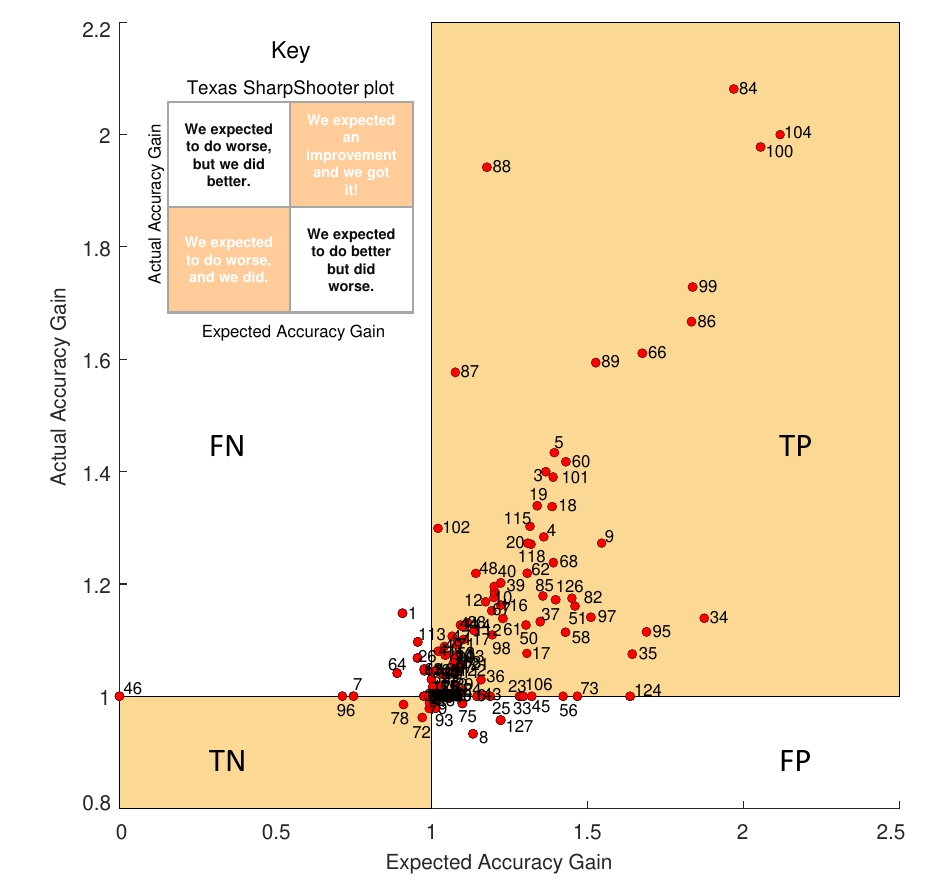}
\caption{
Comparison of Euclidean distance versus constrained DTW for 128 data sets. In the Texas Sharpshooter plot, each data set falls into one of four possibilities corresponding to four quadrants. We optimize the performance of DTW by learning a suitable warping window width and compare the expected improvement with the actual improvement. The results are strongly supportive of the claim that DTW is better than Euclidean distance for most problems. Note that some of the numbers are hard to read because they overlap. 
}
\label{fig:UCRArchive-p7}
\end{figure}

In Fig.~\ref{fig:UCRArchive-p7} we use the Texas Sharpshooter plot popularized by Batista et al. \cite{Batista2014} to show the baseline result comparison between 1-NN ED and 1-NN constrained DTW on 128 data sets of the new archive. 
 
The Texas Sharpshooter plot is introduced to avoid the Texas sharpshooter fallacy \cite{Batista2014}, that is a simple logic error that seems pervasive in time series classification papers. Many papers show that their algorithm/distance measure are better than the baselines/competitors on some data sets, ties on many and loses on some. They then claim their method works for some domains and thus it has value. However, it is not useful to have an algorithm that are good for some problems unless you can tell \textit{in advance} which problems they are. The Texas Sharpshooter plot in Fig.~\ref{fig:UCRArchive-p7} compares ED and constrained DTW distance by showing the expected accuracy gain (based solely on train data) versus the actual accuracy gain (based solely on test data) of the two methods. Note that here, the improvement of constrained DTW over ED is almost tautological, as constrained DTW subsumes ED as a special case. More generally, these plots visually summarize the strengths and weaknesses of rival methods.

\subsection{Some Notes on the Old Archive}

We reverse the train/test split of fourteen data sets to make them consistent with their original release, i.e. when they were donated to the archive, according to the wish of the original donors. These data sets were accidentally reversed during the 2015 expansion of the archive. The train/test split of these data set now agree with the train/test split hosted at the UEA Archive \cite{Bagnall2018}, and are the split that was used in a recent influential survey paper titled \textit{``The great time series classification bake off: a review and experimental evaluation of recent algorithmic advances"} \cite{Bagnall2017}. We list these data sets in Table \ref{tab:UCRArchive_table_1}.

\begin{table}[htb]
    \centering
    \caption{Fourteen data sets that have the train/test split reversed for the new archive expansion}
    \begin{tabular}{ l l }  
        \toprule
        Data set name &   \\
        \midrule
        DistalPhalanxOutlineAgeGroup    &	DistalPhalanxOutlineCorrect \\
        DistalPhalanxTW &   Earthquakes \\
        FordA   &	FordB \\
        HandOutlines    &	MiddlePhalanxOutlineAgeGroup \\
        MiddlePhalanxOutlineAgeCorrect  &	MiddlePhalanxTW \\
        ProximalPhalanxTW   &	Strawberry \\
        Worms   &	WormsTwoClass \\
        \bottomrule
    \end{tabular}
    \label{tab:UCRArchive_table_1}
\end{table}

Among the 85 data sets of the old archive, there are twelve data sets that at least one algorithm gets 100\% accuracy \cite{Lines2018, Bagnall2018}. We list them in Table \ref{tab:UCRArchive_table_2}.
 
 \begin{table}[htb]
    \centering
    \caption{Twelve ``solved" data sets, which at least one algorithm gets 100\% accuracy}
    \begin{tabular}{ l l l l}  
        \toprule
        Type    &   Data set name    &   Type    &	 Data set name \\
        \midrule
        Image   &   BirdChicken &	Sensor  &	Plane \\
        Spectrograph    &	Coffee  &	Simulated   &   ShapeletSim \\
        ECG &	ECGFiveDays &	Simulated   &	SyntheticControl \\
        Image   &	FaceFour    &	Sensor  &	Trace \\
        Motion  &   GunPoint    &	Simulated   &       TwoPatterns \\
        Spectrograph    &	Meat    &	Sensor  &	Wafer \\
        \bottomrule
    \end{tabular}
    \label{tab:UCRArchive_table_2}
\end{table}

\subsection{Data Set Highlights}

\subsubsection{\textit{GunPoint} data sets}

\begin{figure}[htb]
\centering
\includegraphics[page=8,trim={10cm 4.7cm 10cm 4.7cm}, clip, width=0.95\columnwidth]{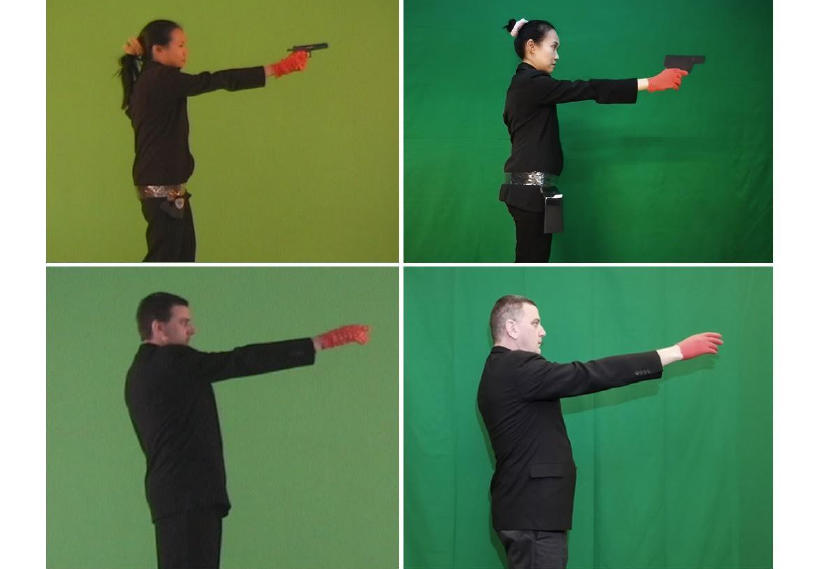}
\caption{
\textit{left}) \textit{GunPoint} recording of 2003 and \textit{right}) \textit{GunPoint} recording of 2018. The female and male actors are the same individuals recorded fifteen years apart. 
}
\label{fig:UCRArchive-p8}
\end{figure}

The original \textit{GunPoint} data set was created by current authors Ratanamahatana and Keogh in 2003. Since then, it has become the ``iris data" of the time series community \cite{Fisher1936}, being used in over one thousand papers, with images from the data set appearing in dozens of papers (see Fig.~\ref{fig:UCRArchive-p8}). As part of these new release of the UCR Archive, we decided to revisit this problem, by asking the two original actors to recreate the data.   

We record two scenarios, ``Gun" and ``Point". In each scenario, the actors aim at a target at eye level before them. We strived to reproduce in every aspect the recording of the original \textit{GunPoint} data set created 15 years ago. The difference between Gun and Point is that in the Gun scenario, the actor holds a replica gun. They point the gun at the target, return the gun back to the waist holster and then brings their free hand to a rest position to complete an action. Each complete action conforms to a five-second cycle. We filmed with a commodity smart-phone Samsung Galaxy 8. With 30fps, this translates into 150 frames per action. We generated a time series for each action by taking the x-axis location of the centroid of the red gloved hand (see Fig.~\ref{fig:UCRArchive-p8}). We merged the data of this new recording with the old \textit{GunPoint} data to make several new data sets.

The data collected now spans two actors {F,M}, two behaviors {G,P}, and two years {03,18}. The task of the original \textit{GunPoint} data set was differentiating between the Gun and the Point action: {FG03, MG03} vs. {FP03, MP03}. We have created three new data sets. Each data set has two classes; each class is highly polymorphic with four variants characterizing it.

\begin{itemize}
    \item 
	\textit{GunPointAgeSpan}: {FG03, MG03, FG18, MG18} vs. {FP03, MP03, FP18, MP18}. The task is to recognize the actions with invariance to the actor, as with \textit{GunPoint} before, but also be invariant to the year of recording.
	\item
	\textit{GunPointOldVersusYoung}: {FG03, MG03, FP03, MP03} vs. {FG18, MG18, FP18, MP18}, which asks if a classifier can detect the difference between the recording sessions due to (perhaps) the actors aging, differences in equipment and processing; though as noted above, we tried to minimize such inconsistencies. In this case, the classifier needs to ignore the action and actor.
	\item
	\textit{GunPointMaleVersusFemale}: {FG03, FP03, FG18, FP18} vs. {MG03, MP03, MG18, MP18}, which asks if a classifier can differentiate between the build and posture of the two actors.  
\end{itemize}

\subsubsection{\textit{GesturePebble} data sets}

The archive expansion includes several data sets whose time series exemplars can be of different lengths. The \textit{GesturePebble} data set is one of them. For ease of data handling, we pad enough NaNs to the end of each time series, to make it the same length of the longest time series. Some algorithms/distance measures can handle variable-length data directly; other researchers may have to process such data by truncation or re-normalization etc. We deliberately refrain from offering any advice on how to best do this. 

\begin{figure}[htb]
\centering
\includegraphics[page=9,trim={10cm 7cm 10cm 7cm}, clip, width=1\columnwidth]{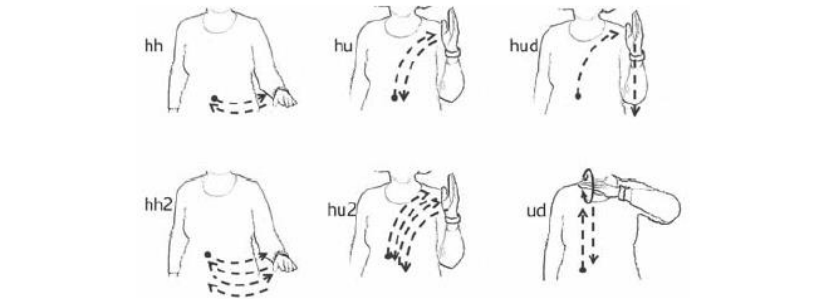}
\caption{
The dot marks the start of a gesture. The labels (hh, hu, hud, etc) are used by original authors of the data and may not have any special meaning. The gestures are selected based on criteria that they are characterized by the wrist movements; they simple and natural enough to replicate; and they can be related to commands to control devices \cite{Mezari2017}. 
}
\label{fig:UCRArchive-p9}
\end{figure}

\begin{figure}[htb]
\centering
\includegraphics[page=10,trim={5.4cm 6.7cm 5.4cm 6.7cm}, clip, width=1\columnwidth]{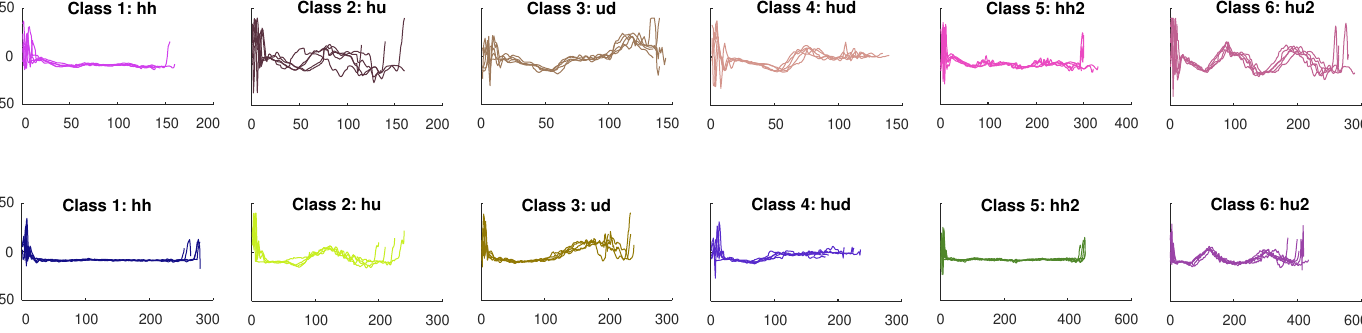}
\caption{
Data from recording of two different subjects performing a same set of six gestures (top row and bottom row). The endpoints of the time series contain the ``tap event", which are abrupt movements to signal the start and end of the gesture (deliberately performed by the subject). The accelerometer data is also under influence of gravity and the device’s orientation during the movement.
}
\label{fig:UCRArchive-p10}
\end{figure}

The \textit{GesturePebble} data set comes from the paper \textit{``Gesture Recognition using Symbolic Aggregate Approximation and Dynamic Time Warping on Motion Data"} \cite{Mezari2017}. This work is among the many that study the application of commodity smart devices as a motion sensor for gesture recognition. The data is collected with the 3-axis accelerometer Pebble smart watch mounted to the participants’ wrist. Each subject is instructed to perform six gestures depicted in Fig.~\ref{fig:UCRArchive-p9}. The data collection included four participants, each of which repeated the gesture set in two separate sessions a few days part. In total, there are eight recordings, which contain 304 gestures. Since the duration of each gestures varies, the time series representing each gesture are of different lengths. Fig.~\ref{fig:UCRArchive-p10} shows some samples of the data.

We created two data sets from the original data, both using only the z-axis reading (out of the three channels/attributes available).

\begin{itemize}
    \item 
	\textit{GesturePebbleZ1}: The train set consists of data of all subjects collected in the first session. The test set consists of all data collected in the second session. This way, data of each subject appear in both train and test set.  
	\item
	\textit{GesturePebbleZ2}: The train set consists of data of two subjects and the test set consists of data of the other two subjects. This data set is intended to be more difficult than \textit{GesturePebbleZ1} because the subjects in the test set do not appear in the train set (presumably that each participant possesses unique gait and posture and move differently). Baseline results confirm this speculation.
\end{itemize}

\subsubsection{EthanolLevel data set}

\begin{figure}[htb]
\centering
\includegraphics[page=11,trim={9cm 7.5cm 9cm 7.5cm}, clip, width=0.95\columnwidth]{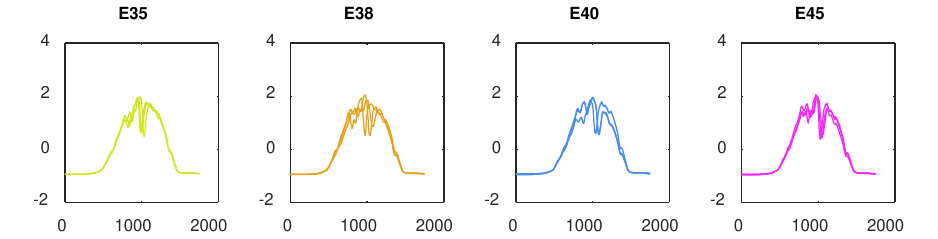}
\caption{
Three examples per class of \textit{EthanolLevel} data set. The four classes correspond to levels of alcohol: 35\%, 38\%, 40\% and 45\%. Each series is 1751 data-point long.
}
\label{fig:UCRArchive-p11}
\end{figure}

This data set was produced as part of a project with Scotch Whisky Research Institute into non-intrusively detecting forged spirits (counterfeiting whiskey is an increasingly lucrative crime). One such candidate method of detecting forgeries is by examining the ethanol level extracted from a spectrograph. The data set covers twenty different bottle types and four levels of alcohol: 35\%, 38\%, 40\% and 45\%. Each series is a spectrograph of 1,751 observations. Fig.~\ref{fig:UCRArchive-p11} shows some examples of each class.

This data set is an example of when it is wrong to merge and resample, because the train/test sets are constructed so that the same bottle type is never in both data sets. The data set was introduced in \textit{``HIVE-COTE: The hierarchical vote collective of transformation-based ensembles for time series classification"} (Lines, Taylor, and Bagnall 2016).

\subsubsection{\textit{InternalBleeding} data sets} 

\begin{figure}[htb]
\centering
\includegraphics[page=12,trim={4cm 5cm 4cm 5cm}, clip, width=1\columnwidth]{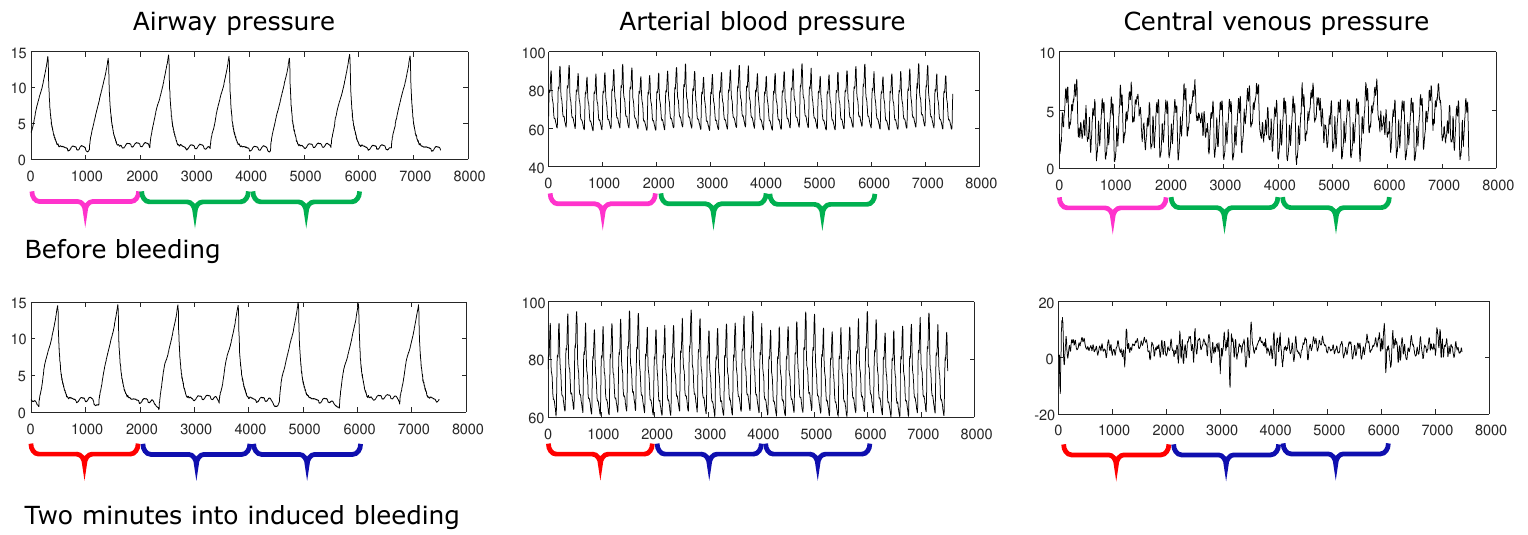}
\caption{
\textit{InternalBleeding} data sets. Class $i$ is the $i^{th}$ individual pig. In the training set, class $i$ is represented by two examples, the first 2000 data points of the \textit{``before"} time series (pink braces), and the first 2000 data points of the \textit{``after"} time series (red braces). In the test set, class $i$ is represented by four examples, the second and third 2000 data points of the \textit{``before"} time series (green braces), and the second and third 2000 data points of the \textit{``after"} time series (blue braces).
}
\label{fig:UCRArchive-p12}
\end{figure}

The source data set is data from fifty-two pigs having three vital signs monitored, before and after an induced injury \cite{Guillame-Bert2017}. We created three data sets out of this source: \textit{AirwayPressure} (airway pressure measurements), \textit{ArtPressure} (arterial blood pressure measurements) and \textit{CVP} (central venous pressure measurements). Fig.~\ref{fig:UCRArchive-p12} shows a sample of these data sets. In a handful of cases, data may be missing or corrupt; we have done nothing to rectify this. 

These data sets are interesting and challenging for several reasons. Firstly, the data are not phase-aligned, which may call for a phase-invariant or an elastic distance measure. Secondly, the number of classes is huge, especially relative to the number of training instances. Finally, each class is polymorphic, having exactly one example from the pig when it was healthy, and one from when it was injured. 

\subsubsection{Electrical Load Measurement data - Freezer data sets}

This data set was derived from a multi-institution project entitled Personalized Retrofit Decision Support Tools for UK Homes using Smart Home Technology (REFIT) \cite{Murray2015}. The data set includes data from twenty households from the Loughborough area over the period of 2013-2014. All data are from freezers in House 1. This data set has two classes, one representing the power demand of the fridge freezer in the kitchen, the other representing the power demand of the (less frequently used) freezer in the garage.

\begin{figure}[htb]
\centering
\includegraphics[page=13,trim={10.5cm 6.9cm 10.5cm 6.9cm}, clip, width=0.95\columnwidth]{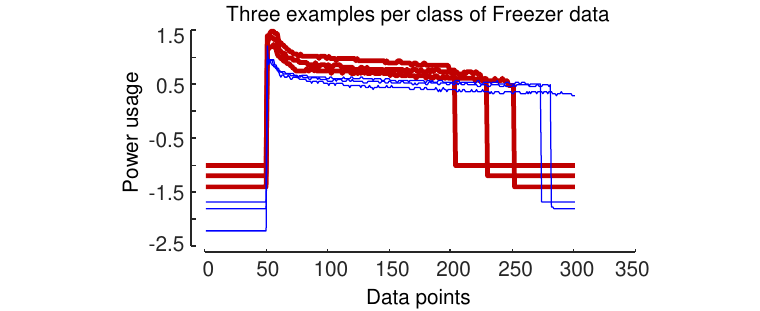}
\caption{
Some examples from the two-class \textit{Freezer} data set. The two classes (\textcolor{blue}{blue}/fine lines vs. \textcolor{red_boston}{red}/bold lines) represent the power demand of the fridge freezers sitting in different locations of the house. The classes are difficult to tell apart globally but they differ locally.
}
\label{fig:UCRArchive-p13}
\end{figure}

The two classes are difficult to tell apart globally. Each consists of a flat region (the compressor is off), followed by an instantaneous increase (the compressor is switched on), followed by a slower decease as the compressor builds some rotational inertial. Finally, once the temperature has been lowered enough, there is an instantaneous fall back to a flat region (this part may be missing in some exemplars). The amount of time the compressor is on can vary greatly, but that is not class-dependent. In Fig.~\ref{fig:UCRArchive-p13} however, if you examine the region just after the fiftieth data point, you can see a subtle class-conserved difference in how fast the compressor builds rotational inertial and decreases its power demand. An algorithm that can exploit this difference could do well on these data sets; however, global algorithms, such as 1-NN with Euclidean distance may have a hard time beating the default rate. 

\textit{Freezer} is an example of data sets with different train/test splits (the others are \textit{InsectEPG} and \textit{MixedShapes}). We created two train set versions: a smaller train set and a regular train set (both are accompanied by a same test set). This is to meet the community's demand of  benchmarking algorithms that are able to work with little train data, for example, generating synthetic time series to augment sparse data sets \cite{Dau2018LearningW, Forestier2017}. Some algorithms produce favorable results when training exemplars are abundant but deteriorate when the train data are scarce. 

\section{Conclusions and Future Work \label{conclusions}}

We have introduced the 128-data set version of the UCR Time Series Archive. This resource is made freely available at the online repository in perpetuity \cite{Dau2018UCRArchivePage}. A separate web page in supporting of this paper is also available \cite{Dau2019UCRArchivePaperWebpage}. 
We have further offered advice to the community on best practices on using the archive to test classification algorithms, although we recognize that the community is free to ignore all such advice. Finally, we offered a cautionary tale about how easily practitioners can inadvertently mis-attribute improvements in classification accuracy. We hope this will encourage all users (including the current authors) to have deeper introspection about the evaluation of proposed distance measures and algorithms.  



\balance
\bibliographystyle{IEEEtran}
\bibliography{UCRArchive2018}

\end{document}